\pdfoutput=1

\documentclass[11pt]{article}

\usepackage[]{acl}

\usepackage{times}
\usepackage{latexsym}
\usepackage{amsmath}
\usepackage{amssymb}
\usepackage{graphicx}

\usepackage{booktabs}
\usepackage[T1]{fontenc}

\usepackage[utf8]{inputenc}

\usepackage{microtype}

\usepackage[disable]{todonotes}

\title{When can I Speak? Predicting initiation points for spoken dialogue agents}

\newcommand{\Iks}{\ensuremath{I^k_{\text{spkr}}}}
\newcommand{\Ikt}{\ensuremath{I^k_{\text{tgr}}}}

\newcommand{\Ic}[1]{\ensuremath{I^{#1}_{\text{cur}}}}
\newcommand{\It}[1]{\ensuremath{I^{#1}_{\text{tgt}}}}
\newcommand{\maep}{Mean Absolute Error w.r.t.\ Predicted Lead Time\ }
\newcommand{\maet}{Mean Absolute Error w.r.t.\ True Lead Time\ }
\newcommand{\wrt}{w.r.t.\ }
\newcommand{\MMT}{\ensuremath{\text{MMAE-True}(-0.5, 1)}}
\newcommand{\MMP}{\ensuremath{\text{MMAE-Pred}(0, 1)}}

\author{Siyan Li \\
  \\
  \\\And
  Ashwin Paranjape \\
  Stanford University \\
  \texttt{\{siyanli, ashwinp, manning\}@cs.stanford.edu} \\\And 
  Christopher D. Manning \\
  \\}

\begin{document}
\maketitle
\begin{abstract}
Current spoken dialogue systems initiate their turns after a long period of silence (700-1000ms), which leads to little real-time feedback, sluggish responses, and an overall stilted conversational flow. Humans typically respond within 200ms and successfully predicting initiation points in advance would allow spoken dialogue agents to do the same. In this work, we predict the lead-time to initiation using prosodic features from a pre-trained speech representation model (wav2vec 1.0) operating on user audio and word features from a pre-trained language model (GPT-2) operating on incremental transcriptions. To evaluate errors, we propose two metrics \wrt predicted and true lead times. We train and evaluate the models on the Switchboard Corpus and find that our method outperforms features from prior work on both metrics and vastly outperforms the common approach of waiting for 700ms of silence.
\end{abstract}

\section{Introduction}
Spoken dialogue agents have exploded in popular use (e.g., Alexa, Siri, and Google Home). 
However, they only support explicit turn-taking mechanisms: they detect user initiation and barge-ins using wake-words and identify end of user turns based on a silence period (typically between 700--1000ms).
Turn-taking feels unnatural under such mechanisms, leading to less ``conversational'' interactions \cite{woodruff2003push}. 
This is particularly damaging for open-ended social conversations where thoughtful silences get wrongly interrupted \cite{chi2021neural}.
To fix this issue, we predict initiation opportunities for spoken dialogue agents for both turn-taking and backchanneling. 

\begin{figure}
    \centering
    \includegraphics[width=\linewidth]{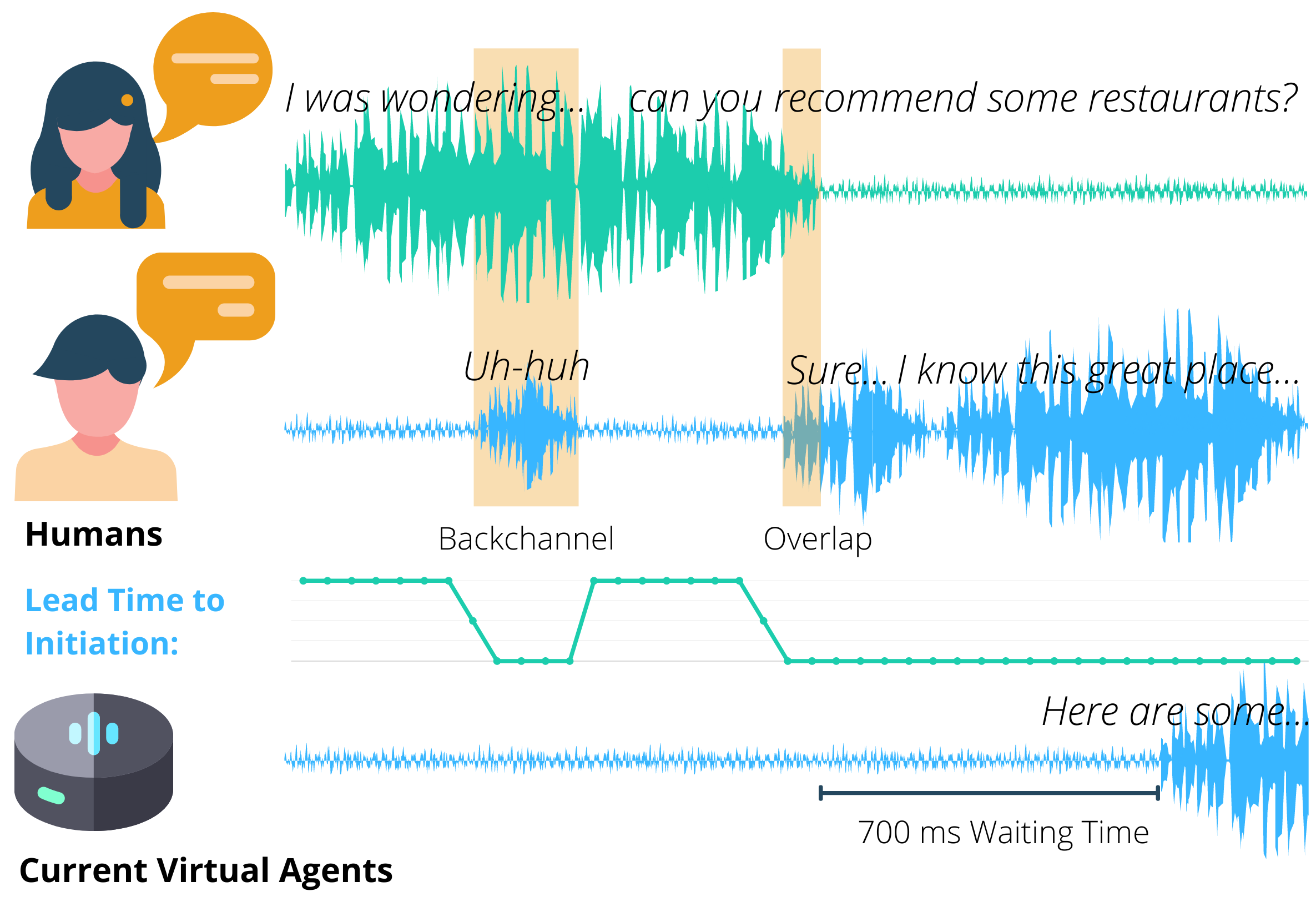}
    \caption{Humans produce overlapping speech with small gaps. By predicting lead to initiation, virtual agents can respond without long waiting periods}
    \label{fig:my_label}
\end{figure}
Prior work predicting initiation points uses prosodic features like pitch and frequency variation with bag-of-embeddings to predict backchannels \cite{ruede-enhance} and turn-completion \cite{skantze2017towards}, and more recently, \citet{ekstedt-skantze-2021-projection} finetuned GPT-2 on dialogue datasets to predict turn-completion using only word features. 
However, they either predict a binary label indicating initiation in a wide event horizon, which is imprecise; or they predict a binary label for an initiation to happen at a set offset in the future, in which case a single incorrect prediction leads to a missed initiation. 

As a robust generalization of previous approaches, we predict the lead time to initiation as a continuous value. 
We model initiation (next utterance from a different speaker) directly and not end-of-turn because there is a variable (and possibly negative) gap between the two \citep{skantze2021turn}. 
In this work, we combine two models: wav2vec 1.0 \cite{schneider2019wav2vec} for representing prosodic features and finetuned GPT-2 \cite{radford2019language} for word features.
We model the task with a Gaussian Mixture Model (GMM) to account for inherent uncertainty. We train and evaluate our models on Switchboard \cite{switchboard} and find that the combination of the pretrained models performs the best, vastly outperforming a silence-based baseline that waits for 700ms of silence and baselines using features from prior work.

\section{Related Work}
Prior work for dialogue turn-taking either uses silent gaps as cues or predicts future events repeatedly. A key issue with systems that use silent gaps as initiation cues \citep{virtualrapport,cohen2004voice,witt2015modeling} is the difficulty of adjusting the silence thresholds to accommodate dialogue states \citep{skantze2021turn}. When predicting turn-taking repeatedly, i.e. predicting future actions at every timestep, acoustic features such as pitch and frequency are often used, with additional linguistic features including part-of-speech or word embeddings \citep{ruede-enhance,Ruede0SW17,skantze2017towards,ward2018turn,roddy2018investigating}. More recently, \citet{ekstedt-skantze-2021-projection} implement a spoken dialogue system for travel conversations using TurnGPT \cite{ekstedt-skantze-2020-turngpt}. 
However, a short silence threshold is still used to determine initiation of agent responses.

Outside of dialogue, \citet{neumann2019future} propose probabilistic models for predicting events in videos, \citet {lei-etal-2020-likely} forecast frames and \citet{vondrick2016anticipating} forecast actions. Time-to-event analysis in the medical domain involves modeling patient status as a function of time \citep{multi-state-tte,soleimani2017scalable}. 



\section{Methods}

\subsection{Setup}
$\mathbf{\Iks}$ is the time of $k$-th initiation (both backchannels and transitions) by a speaker. 
We use the \textbf{current speaker}'s audio and transcript information to predict the the \textbf{lead time to initiation}, $\hat{\tau}_t$, of the \textbf{target speaker}. 
When the current speaker is speaking, we consider an \textbf{event horizon} $\delta_{max}$ to narrow the prediction range and at time $t$, define the true \textbf{lead time to initiation} as $\tau_t =\min(\delta_{max}, \Ikt - t)$. 
When the target speaker is speaking, we set $\tau_t = 0$, to ensure a  well-balanced distribution. 

\subsection{Models}

We make two novel contributions. First, we fuse rich contextual prosodic features from a pretrained wav2vec model with contextual word representations from a pretrained GPT-2 model. Prior work has not used such rich contextual prosodic features nor their combination with word representations.  
Second, prior work does not model the inherent uncertainty of initiations.
Inspired by the video event prediction literature \cite{neumann2019future}, we do this using a Gaussian mixture model and maximize model likelihood under the data distribution.

\begin{figure*}
    \centering
    \includegraphics[width=\linewidth]{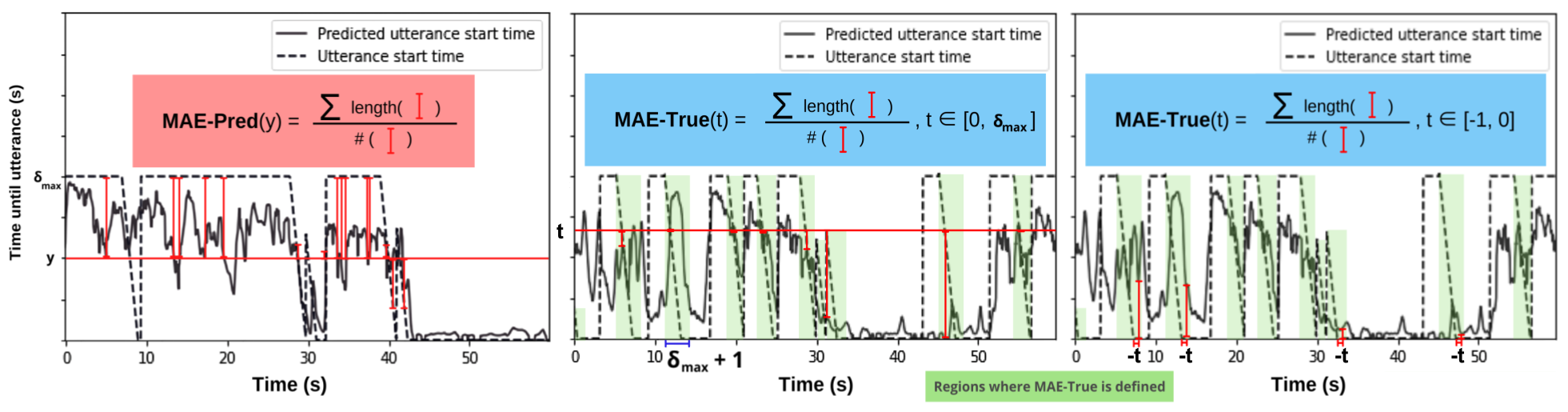}
    \caption{An explanation of our metrics. The red vertical intervals correspond to $|\tau_x - \hat{\tau}_x|$ in the equations. As illustrated, MAE-Pred(y) evaluates the expected error when a model predicts value y. For MAE-True(t), we highlight the regions where MAE-True can be calculated in green; depending on how long the current speaker's next utterance is, the region has a maximum length of $\delta_{max} + 1$.}
    \label{fig:my_label}
\end{figure*}
\subsubsection{Features}
Features are extracted from the current speaker's voice channel and transcript. We suffix model names with abbreviated versions of the features they use. 

\textbf{Wav2vec Embeddings (W):} Raw audio is fed into Wav2vec 1.0 \cite{schneider2019wav2vec} to obtain convolutional embeddings. We choose Wav2vec 1.0 because of its unidirectional nature, which enables handling efficient incremental processing of audio. We keep the model weights frozen. 

\textbf{GPT-2 Embeddings (G):} This is the GPT-2 Small \citep{radford2019language} embedding of the last salient word from the target speaker after feeding in prior utterances. The embedding is updated incrementally as more utterances are transcribed. We fine-tune the GPT-2 model during training. 

\textbf{RMSE (R):} We select the Root Mean Square Energy (RMSE) of the raw waveform to signal current speaker silence. It simulates audio energy and power in features from prior work.

\textbf{Additional Prosodic Features (A):} Previous work explores pre-neural prosodic features \citep{ruede-enhance,Ruede0SW17,skantze2017towards}; to compare our approach with previous approaches, we include pitch and frequency, both represented as a number for each frame. The prosodic features, including RMSE, are calculated with a frame shift of 50 ms and a window length of 100 ms. Additional details for feature implementation are in Appendix \ref{app:feats}.

Wav2vec features are subsampled to 50 ms by selecting embeddings at every 50ms and for other audio features by adjusting the frame shift. Audio features are concatenated and input to an LSTM network. When GPT-2 embeddings are used, they are concatenated with the LSTM's final hidden state. 
This is fed into a linear head. More training details are presented in Appendix \ref{app:training}.
\subsubsection{Gaussian Mixture Model}
%
There is an inherent uncertainty in the precise location of an initiation (e.g., it can occur a few milliseconds before or after the prediction) and a single Gaussian is sufficiently powerful to model it because the uncertainty is localized. 
However, a speaker can initiate at many points in time that are far apart, for e.g., at the completions of grammatical clauses that can happen hundreds of milliseconds apart. 
We use a Gaussian mixture model (GMM) to capture this multimodal prediction space.

At every time step, we predict the parameters: mean, variance and weights, for $T$
Gaussian distributions $\{\mu, \sigma, h\}_{[1..T]}$.  
The training objective is to maximize the log of the summed likelihood of $\tau_t$:
$$\log \Big{(}\sum_{i=1}^T h_i \cdot \frac{1}{ \sigma_i \sqrt{2\pi}} \cdot \exp -\frac{(\tau_t - \mu_i)^2}{2 \sigma_i^2} \Big{)}$$
At inference, we use the mean of the Gaussians.

\subsubsection{Baselines}
\textbf{Silence Baseline:} We compare our models with an RMSE-based non-neural baseline. We detect voice activity based on whether RMSE is above a certain threshold (0.01 for this work). If there is a gap of more than 700ms in voice-activity, the baseline predicts an initiation $\tau_t = 0$ at the current time, otherwise predicts $\delta_{max}$. 

\textbf{GMM-AG:} We use this baseline as a proxy for \citet{ruede-enhance}, where pitch, power, and FFV are used as the prosodic features, and word2vec embedding of the most recent salient word is the linguistic feature. We simulate these features using RMSE, pitch and frequency (the prosodic features), and GPT-2 embeddings. 

\textbf{GMM-G:} \citet{ekstedt-skantze-2020-turngpt} use GPT-2 to emulate possible continuations of the current conversation in order to decide turn-relevant places. Although we do not use the same algorithm, we still use GPT-2 embedding as a feature. We train a GMM on last-salient-word GPT-2 embeddings only, and use this as a representative baseline for \citet{ekstedt-skantze-2020-turngpt}.

\textbf{GMM-WGR-1:} We train a Gaussian mixture model with $T=1$ Gaussian to examine whether using multiple Gaussian models to capture different factors for utterance timing is necessary. This model is trained on the same data as our GMM-WGR model, with Wav2vec, GPT-2, and RMS features.

\subsection{Training and Evaluation Data}
For training, we randomly sample 60 second audio segments that have its first target speaker initiation in the first 5 to 10 seconds. 
This is to make sure that there is at least one initiation with enough context. We backpropagate losses only in a limited range around each initiation \It{i}, $[\It{i} - 2\delta_{max}, \It{i} + 1]$ 
This is to ensure a balanced distribution of $\tau_t$. For evaluation and testing, we instead cover entire dialogues by collecting 60-second segments every 20 seconds.
We randomly choose the target speaker for each segment.
\begin{figure*}[!th]
    \centering
    \includegraphics[width=\textwidth]{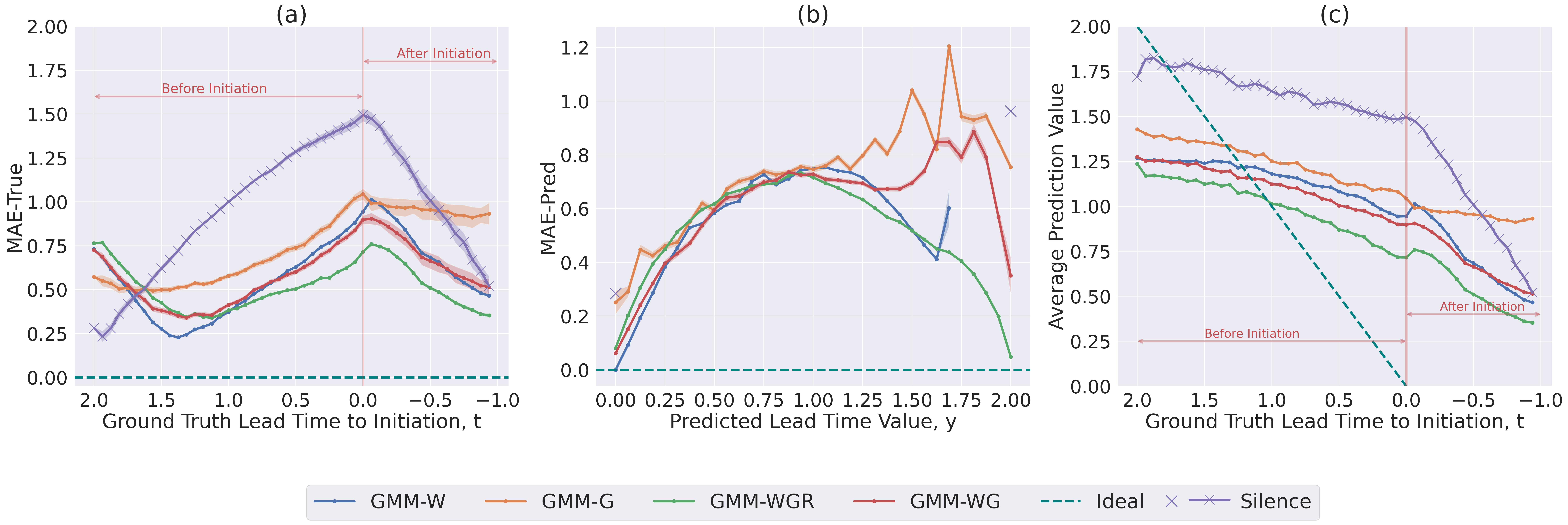}
    \caption{(a) MAE-True, (b) MAE-Pred, and (c) average predicted lead time values for representative neural models and the silence baseline. 95\% C.I.\ are represented by the lightly shaded regions. A perfect model would achieve the ``ideal'' (dashed) lines. In (b), because the silence based model only predicts $0$ or $\delta_{max}$, only these two points are defined in plot (b) for the silence based baseline. The corresponding MAE-Pred values for the silence baseline are indicated as crosses in plot(b). All of our models, including the best performing GMM-WGR, significantly outperform the silence-based model that waits for 700 ms.}
    \label{fig:results}
\end{figure*}

\subsection{Metrics}
To measure the performance of our models that produce continuous values, previous work's classification-based metrics are insufficient to differentiate between a prediction error of 0.2 versus 2 seconds. Additionally, we want to differentiate between how precise model predictions are and how well they cover the initiations observed in the dataset. 
We improve upon Time-to-event error from \citet{neumann2019future}, and propose \maep (MAE-Pred) and \maet (MAE-True) as analogues of precision and recall that improve existing metrics \cite{skantze2017towards}.
If a practitioner needs $l$ seconds to generate a response, MAE-Pred($l$) gives the expected error when the model predicts $l$ (precision) and MAE-True($l$) gives the expected error with the true lead time is $l$ (recall). With the set $S$ representing the timesteps included in the calculations, both metrics can be represented as
$${\sum\nolimits_{x \in S}|\tau_x - \hat{\tau}_x|}/{|S|}$$
Specifically, for \textbf{MAE-Pred($y$)}:
$$S = \{x | \hat{\tau}_x = y\}, y \in [0, \delta_{max}]$$
For \textbf{MAE-True($t$)}:
$$
    S = \{\It{i} - t \}, t \in [-1, \delta_{max}] \cap [\It{i}-\Ic{j+1}, \It{i} - \Ic{j}]
$$
for all target-speaker initiations \It{i}, limiting to intervals between two consecutive initiations by the current speaker. When $t \leq 0$, the initiation has already occurred and $\tau_t = 0$. We quantize both true and predicted values into 16 buckets per second.

As an aggregated metric, we propose Macro-MAE (MMAE). We define MMAE-X($a, b$) $= \sum_{v \in S_{ab}}$MAE-X$(v) / |S_{ab}|$, where $S_{ab}$ is the set of bucket values between $a$ and $b$ for a given set $S$. 
We define $1$ second before and $0.5s$ after initiation as the interval of interest for MMAE-True, and similarly predicted values between $0$ and $1$ for MMAE-Pred. We compute $\text{MMAE} = \MMT + \MMP$ as a single number quantifying model performance.




\section{Experiments}
For training and evaluation, we use audio conversations from Switchboard \citep{switchboard}. 
We select a random set of 200 training, 20 validation, and 20 test dialogues out of a total of 1000 dialogues due to computational constraints. 
We use the validation set to select the best performing checkpoint based on MMAE scores and report the numbers on the test set. For the GMM models, we experimented with $T = 1, 5, 10, 15, 20$, and found $T=15$ to be the best-performing. 
\footnote{Our code for the models and for training is available at \url{https://github.com/siyan-sylvia-li/icarus_final}} 

We plot the MAE-Pred and MAE-True values in Figure~\ref{fig:results} and the show the MMAE values in Table~\ref{tab:num_results}. 
A perfect model would have $0$ error. As a diagnostic tool, we also plot the average prediction for each $t$ used in MAE-True (Figure~\ref{fig:results} (c)). 
Here, we expect a perfect model to be a line with a slope of $-1$ passing through the origin before flattening out at 0. 
We see that for all models MAE-True peaks (roughly) at initiation (Figure~\ref{fig:results} (b))). 
Despite all the cues leading up to an initiation in the data, it is still highly optional and the models aren't able to predict it perfectly. 
Soon afterward, as the target speaker stays silent the models predict smaller lead times to initiation (steeper downward slope in Figure~\ref{fig:results} (c)) and the MAE-True reduces. 
On the other hand, for all trained models (GMM-*), we see that MAE-Pred reduces for smaller values of $y$ (Figure~\ref{fig:results} (c)) indicating that the trained models are very precise when they predict near-term initiations. 

\begin{table}[!h]
    \centering
    \small
    \begin{tabular}{@{}lrrrr@{}}
    \toprule
         & \multicolumn{3}{c}{Eval} & \multicolumn{1}{c}{Test} \\
             \cmidrule(lr){2-4}
             \cmidrule(lr){5-5}
        \textbf{Model} & MT  & MP  & MMAE & MMAE \\
    \midrule
        GMM-AG & 0.90 & 0.63 & 1.53 & 1.51 \\
        GMM-G &0.90 &0.60 & 1.50 & 1.42 \\
        GMM-WGR-1 & 0.67 & 0.59 & 1.26 & 1.30 \\
        Silence* & 1.33 & 0.60 & 1.93 & 1.88 \\
        \midrule
        GMM-W & 0.70 & \textbf{0.49} & 1.19 & 1.22\\
        GMM-WG  & 0.67& 0.51& 1.18 & 1.19 \\
        \textbf{GMM-WGR} & \textbf{0.63} & 0.52 & \textbf{1.15} & \textbf{1.11} \\
        \bottomrule
    \end{tabular}
    \caption{Performance of different models on the evaluation and the test dialogues, as measured Macro-MAE values. MT = \MMT, MP = \MMP. * Since only $0$ and $\delta_{max}$ are valid predictions for Silence Baseline, we use (MAE-Pred(0) + MAE-Pred($\delta_{max}$))/2 as \MMP.}
    \label{tab:num_results}
\end{table}
%
Our models outperform the silence baseline by a large margin in most time windows prior to and after initiations (Figure~\ref{fig:results} and Table~\ref{tab:num_results}). GMM-WGR outperforms prior work baselines: GMM-G (TurnGPT) and GMM-AG (\citet{ruede-enhance}).
%

Comparing GMM-WG vs.\ GMM-G, Wav2vec features reduce MAE-True after initiation and stabilizes MAE-Pred for small predicted lead times; GMM-G's predictions stay constant after initiations, because it can only access the transcript from the current speaker. Comparing GMM-WG vs.\ GMM-W, GPT-2 features reduce MAE-True near initiations, possibly because they provide the model with word cues. GMM-WGR has a lower MMAE-True(-0.5, 1) compared to GMM-WG, indicating that Wav2vec doesn't capture silences as well as RMSE.
GMM-WGR-1, our baseline with one Gaussian, performs poorly compared to GMM-WGR, highlighting the importance of the Gaussian mixture. 


\section{Conclusion}

We present the task of lead time to initiation prediction as a continuous-valued problem, collapsing transition and backchannel timing problems into one. We additionally propose metrics to capture precision and coverage in these predictions. Our models trained on pretrained prosodic and verbal embeddings consistently outperform the commonly-used silence baseline. We believe our work will build a foundation for more naturalistic virtual agents with human-like conversational behaviors.





\bibliography{anthology,custom}
\bibliographystyle{acl_natbib}

\appendix

\section{Appendix}
\begin{figure*}[!h]
    \centering
    \includegraphics[width=\textwidth]{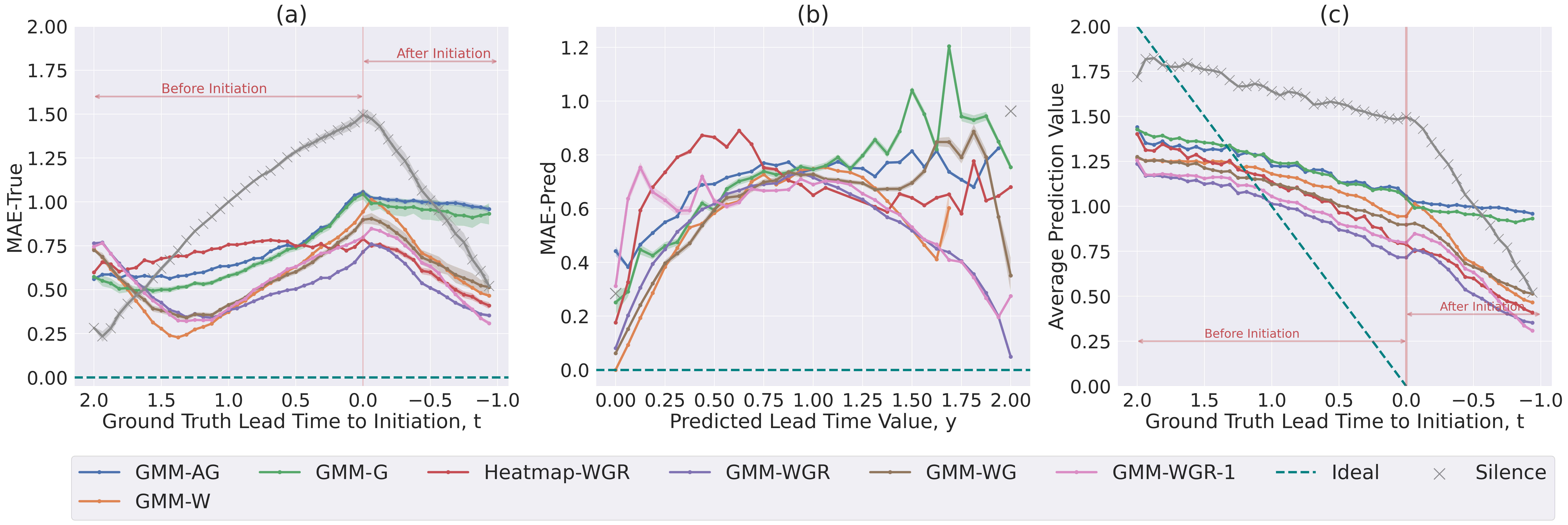}
    \caption{MAE-True, MAE-Pred graphs for all trained models. We also include the graph of average predicted lead time values given true lead time to initiation.}
    \label{fig:my_label}
\end{figure*}

\label{sec:appendix}
\subsection{Feature Implementation} \label{app:feats}
\begin{enumerate}
    \item Pitch: \url{https://pytorch.org/audio/main/functional.html\#compute-kaldi-pitch}
    \item Frequency: \url{https://librosa.org/doc/main/generated/librosa.yin.html}
    \item Root Mean Square Energy: \url{https://librosa.org/doc/main/generated/librosa.feature.rms.html}
\end{enumerate}

\subsection{Training Details} \label{app:training}
The models are trained on one A100 GPU. All model LSTM's have two layers with 128 hidden units. Each epoch approximately last 1000 seconds, and we train each neural model for 7 epochs, at which point overfitting would have definitely occurred. We train all models with dropout 0.1, Adam optimizer, and a weight decay of 0.0001. We include a comprehensive list of our models and their training details in Table~\ref{tab:trained_stats}.

\subsection{Additional Model: Heuristic Heatmap}

We have tried training another probabilistic model from \citet{neumann2019future}, Heuristic Heatmap. We did not find this model to significantly outperform our GMM-Full model, although it does exhibit interesting qualities.

\textbf{Heuristic Heatmap (Histogram-based Density Estimator):} This model captures temporal shifts in the probability distribution of lead time; as the current speaker keeps speaking, the likelihood of an imminent initiation increases for the target speaker, shifting the probability mass from higher to lower lead time values. At every time step, the model produces a probability distribution with $2\delta_{max}r$ ($r = 16$, the resolution of our estimates) bucket values $h_i = P(\tau_t = \frac{2 \delta_{max} i}{2\delta_{max}r})$. Training minimizes the difference between the predicted distribution and a Gaussian
centered at $\tau_t$. During inference, the prediction bucket with the highest probability is returned.

\begin{table}[!th]
    \centering
    \begin{tabular}{lcccc}
    \hline
        \textbf{Model} & \textbf{W} & \textbf{G} & \textbf{Ac} & \textbf{R} \\
    \hline
        GMM-AG &  & \checkmark & \checkmark & \checkmark \\
        GMM-G &  &  \checkmark & & \\
        GMM-W & \checkmark & & & \\
        GMM-WG & \checkmark & \checkmark & & \\
        GMM-WGR& \checkmark & \checkmark & & \checkmark \\
        Heatmap-WGR& \checkmark & \checkmark & & \checkmark \\
        GMM-WGR-1& \checkmark & \checkmark & & \checkmark \\
        \hline
    \end{tabular}
    \caption{The trained models and their features. \textbf{W} represents Wav2vec features, \textbf{G} GPT-2 embeddings, \textbf{Ac} the set of acoustic features (pitch and frequency), \textbf{R} the RMSE of the current speaker waveform.}
    \label{tab:models_app}
\end{table}

\begin{table}[!tbhp]
    \centering
    \small
    \begin{tabular}{lcccc}
    \hline
        \textbf{Model} & $\textbf{MT}_{\textbf{Eval}}$ & $\textbf{MP}_{\textbf{Eval}}$ & $\sum_{\textbf{Eval}}$ & $\sum_{\textbf{Test}}$ \\
    \hline
        GMM-AG & 0.90 & 0.63 & 1.53 & 1.51 \\
        GMM-G &0.84 &0.58 & 1.50 & 1.42 \\
        GMM-W & 0.70 &0.49 & 1.19 & 1.22\\
        GMM-WG  & 0.67& 0.51& 1.18 & 1.19 \\
        GMM-WGR & 0.63 & 0.52 & 1.15 & 1.11 \\
        Heatmap-WGR & 0.80& 0.68 & 1.48 & 1.44 \\
        GMM-WGR-1 & 0.67 & 0.59 & 1.26 & 1.30 \\
        Silence* & 1.33 & 0.60 & 1.93 & 1.88 \\
        \hline
    \end{tabular}
    \caption{Performance of different models on the evaluation and the test dialogues, as measured by the sum of (1) the average MAE-True($t$) on $t \in [1, -0.5]$ ($\textbf{MT}_{\textbf{Eval}}$ and $\textbf{MT}_{\textbf{Test}}$) and (2) the average MAE-Pred($y$) on $y \in [0, 1]$ ($\textbf{MP}_{\textbf{Eval}}$ and $\textbf{MP}_{\textbf{Test}}$). * For the Silence baseline, since only $0$ and $\delta_{max}$ are valid prediction values, we calculate the average of MAE-Pred(0) and MAE-Pred($\delta_{max}$) as $\textbf{MP}_{\textbf{Eval}}$ and $\textbf{MP}_{\textbf{Test}}$.}
    \label{tab:num_results_app}
\end{table}

\begin{table*}
    \centering
    \begin{tabular}{cccc}
    \hline
        \textbf{Model} & \textbf{Features} & \textbf{Learning Rate} & \textbf{Batch Size}\\
        \hline
        GMM-AG & Acoustic features, GPT-2 & \texttt{1e-4} & 16 \\
        GMM-G& GPT-2 embedding & \texttt{1e-4} & 16\\
        GMM-W & Wav2vec representations & \texttt{1e-4} & 32 \\
        GMM-WG & Wav2vec and GPT-2 & \texttt{1e-5} & 16 \\
        GMM-WGR & Wav2vec, GPT-2, and RMSE & \texttt{1e-5} & 32 \\
        GMM-WGR-1 & Wav2vec, GPT-2, and RMSE & \texttt{1e-5} & 15 \\
        Heatmap-WGR & Wav2vec, GPT-2, and RMSE & \texttt{1e-4} & 32 \\
        \hline
    \end{tabular}
    \caption{Set of trained models.}
    \label{tab:trained_stats}
\end{table*}
\subsection{MAE-True and MAE-Pred on All Models}
We also include the graphs for MAE-True, MAE-Pred, and average predictions per ground truth time to initiation values for all of our models. They are presented in Figure~\ref{fig:my_label}.

\end{document}